\documentclass[runningheads]{llncs}
\usepackage{bm}
\usepackage[T1]{fontenc}
\usepackage{graphicx}
\usepackage{booktabs}
\usepackage{multirow}
\usepackage{amsmath}
\usepackage{xcolor}
\usepackage{cite}
\usepackage{url}
\usepackage{float}
\usepackage[hidelinks]{hyperref}

\begin{document}

\title{Brain Metastases Segmentation for BraTS 2026 Task~1:\texorpdfstring{\\}{ }A Multi-Architecture Comparison}

\titlerunning{Multi-Architecture Approach for Brain Metastases Segmentation}

\author{
  Mahdi Islam\inst{1,2}\orcidID{0009-0009-5978-0471} \and
  Musarrat Tabassum\inst{3}
}
\authorrunning{M. Islam and M. Tabassum}

\institute{
  Center for Computational \& Data Sciences, Independent University, Bangladesh \and
  Department of Electrical and Electronic Engineering, Independent University, Bangladesh \\
  \email{mahdi@iub.edu.bd} \and
  Independent Researcher \\
  \email{tabassummusarrat212@gmail.com}
}

\maketitle

\begin{abstract}
Brain metastases are the most common intracranial malignancy, occurring in roughly 30\% of patients with primary solid tumors and carrying a median survival near 5.9~months~\cite{b_prevalence1,b_survival}. Automated
segmentation is critical for treatment planning and volumetric monitoring, but metastases are frequently small, numerous, and heterogeneous in size within a single patient. We compare a plain nnU-Net baseline, a Residual Encoder Large (ResEncL) variant, region-based training, and a Primus transformer model for BraTS-METS~2026 Task~1, using patient-grouped cross-validation to prevent leakage from the longitudinal UCSD subset.
Primus (label-based) is our strongest individual model by aggregate DSC/NSD, achieving 0.710/0.761 (ET), 0.742/0.785 (TC), 0.683/0.689 (WT), and 0.531/0.436 (RC). ResEncL trails Primus on aggregate DSC/NSD but achieves substantially higher lesion-wise F1 (e.g.\ ET: 0.452~vs.\ 0.052); a probability-averaging ensemble of the two only partially preserves ResEncL's F1
advantage (ET lesion-wise F1: 0.064). We further report three postprocessing and label-reconstruction pitfalls we believe generalize beyond this challenge. Code is available at \href{https://github.com/mahdiislam79/BraTS_METS_2026}{github.com/mahdiislam79/BraTS\_METS\_2026}.

\keywords{Brain Metastases \and Tumor Segmentation \and nnU-Net \and Transformer \and Data Leakage \and Small Lesion Detection}
\end{abstract}

\section{Introduction}
Brain metastases represent the most prevalent intracranial malignancy, arising in an estimated 30\% of patients diagnosed with primary solid cancers including lung, breast, and melanoma~\cite{b_prevalence1,b_prevalence2}. Their
development is associated with substantially increased healthcare burden, approximately \$13{,}000 higher per-patient costs when central nervous system spread occurs~\cite{b_cost}, and a median overall survival of less than 6 months~\cite{b_survival}. Stereotactic radiosurgery (SRS) and whole-brain radiation therapy (WBRT) remain primary treatment modalities, both of which demand precise, reproducible lesion delineation to minimize radiation dose to healthy tissue~\cite{b_treatment}.

Accurate volumetric segmentation across serial MRI examinations is a prerequisite for automated treatment-response monitoring, enabling clinicians to detect progression or regression of individual metastatic lesions without inter-observer variability. The longitudinal structure of the BraTS-METS~2026 dataset~\cite{b_brats23}, particularly the UCSD
subset, which provides multiple imaging timepoints per patient makes it uniquely suited for developing such monitoring pipelines, provided that data splitting is handled to prevent temporal leakage between training and validation folds.

Recent advances in automated brain tumor segmentation have been driven primarily by nnU-Net~\cite{b_nnunet}, a self-configuring framework that remains highly competitive across segmentation benchmarks. Architecture extensions, including Residual Encoder variants~\cite{b_nnunet_revisited}, region-based training for overlapping tumor subregion definitions~\cite{b_nnunet_brats}, and transformer-based models such as SwinUNETR~\cite{b_swinunetr} and Primus~\cite{b_primus}, have demonstrated incremental gains. Multi-architecture ensembling yields the strong reported results in prior BraTS
challenges~\cite{b_astaraki,b_krikorian}, by combining
complementary failure modes across convolutional and transformer-based models. Persistent challenges remain in small and multifocal lesion detection, handling of the rare resection cavity (RC) class, and robustness to extreme class imbalance between metastatic lesion volume and surrounding brain tissue.

We present our approach to BraTS-METS~2026 Task~1, segmenting pre- and post-treatment brain metastases into four subregions: enhancing tumor (ET), non-enhancing tumor core (NETC), surrounding non-enhancing FLAIR
hyperintensity (SNFH), and resection cavity (RC). Our pipeline compares a plain nnU-Net baseline, a Residual Encoder Large (ResEncL) variant, a region-based training variant of Primus, and a Primus transformer model, and combines the two strongest individual models (ResEncL and Primus, both label-based) into a probability-averaging ensemble. Beyond this architectural comparison, we report three findings we believe are useful to other implementers building metastasis segmentation pipelines: (i)~a patient-grouped cross-validation scheme that prevents the temporal data leakage inherent to the longitudinal UCSD subset of BraTS-METS; (ii)~an analysis of a postprocessing failure mode morphological opening silently destroying small lesions near the challenge's scoring threshold that we argue generalizes to other small-lesion segmentation pipelines; and (iii)~identification of a labeling-order assumption in region-based nnU-Net reconstruction that can silently corrupt output classes under non-trivial containment hierarchies, together with a correction procedure requiring no retraining.

\section{Method}
\subsection{Dataset and Preprocessing}
We used the BraTS-METS~2025 Lighthouse dataset~\cite{b_brats25_lighthouse}\footnote{Synapse ID: \texttt{syn74274097}.}, reused for BraTS~2026 Task~1, comprising 1{,}496 training cases from nine contributing institutions: Duke, NCI, Missouri, WashU, Yale, UCSF, Northwestern, UCSD, and Ulm~\cite{b_brats23}. Of these,
1{,}296 cases carry expert annotations and were used for supervised training; the 200~Ulm cases are unannotated and were excluded. Four MRI modalities are provided: native T1~(T1n), contrast-enhanced T1~(T1c), T2-weighted~(T2w, non-mandatory from 2025 onward), and T2-weighted FLAIR (T2f). Ground truth labels follow a five-class scheme: 0~=~background, 1~=~NETC, 2~=~SNFH, 3~=~ET, 4~=~RC. RC~(label~4) is present in approximately 13\% of cases and is retained as a separate class distinct from the whole tumor (WT) hierarchy.

Two cases contained out-of-scheme segmentation labels. One was resolved by applying the official corrected-labels overlay; the other (129~voxels affected) was not covered by that correction and was excluded from training. Excluding this single unresolved case, all remaining 1{,}295 annotated cases were
eligible for conversion; a further, separate set of cases was excluded at conversion time for missing one or more of the four required modalities (T2w in particular is non-mandatory and absent for several cases), following a conservative complete-modality inclusion criterion. Twenty-eight cases with all-zero segmentation masks confirmed as legitimate no-visible-disease UCSD cases per the dataset description were retained as valid negative training examples rather than treated as missing annotations.

Data were converted to nnU-Net raw format using standard nnU-Net
fingerprinting and automatic preprocessing. For the region-based training variant, a second version of the dataset was constructed with overlapping region labels in place of the mutually exclusive per-voxel classes used for the other models.

\subsection{Cross-Validation Splits and Leakage Prevention}
The UCSD subset has a longitudinal structure where each patient contributes multiple imaging timepoints, encoded within the case identifier. Naive random splitting or modulo-based fold assignment allows the same patient to appear in both training and validation within a fold, inflating estimated performance. We instead grouped cases by patient identifier and
applied grouped 5-fold cross-validation (\texttt{GroupKFold}).
The resulting fold assignment was reused identically across all model variants for training and model selection; all reported metrics are official validation-server scores, not CV scores.

\subsection{Architectures}
We trained and evaluated four model configurations.

\noindent\textbf{Plain nnU-Net (Baseline).}
Standard 3D full-resolution nnU-Net~v2~\cite{b_nnunet} with default architecture and plans. We also included an earlier, shorter-trained (200-epoch) checkpoint that predates the postprocessing correction described in Section~\ref{sec:postprocessing} and differs from the 1000-epoch baseline in both training length and postprocessing.

\noindent\textbf{ResEncL.}
Residual Encoder Large variant~\cite{b_nnunet_revisited}, trained on the label-based dataset with training configuration otherwise identical to the baseline.

\begin{sloppypar}
\noindent\textbf{Region-Based Training.}
To handle overlapping region definitions (WT\,=\,\{NETC,\,SNFH,\,ET\}; TC\,=\,\{NETC,\,ET\}), we adopted a region-based training scheme~\cite{b_nnunet_brats} with overlapping labels in place of mutually exclusive classes, applied to Primus. We also trained a region-based ResEncL variant, but given the anomalous results with region-based Primus (Section~\ref{sec:region-order}), did not pursue it further and do not report it.
\end{sloppypar}

\noindent\textbf{Primus.}
A Primus~(PrimusV3S) transformer model~\cite{b_primus}, integrated as an nnU-Net~v2 trainer class, trained in both label-based and region-based configurations. The label-based configuration achieved the strongest aggregate DSC/NSD among individual models (Section~\ref{sec:results}); reconstruction of discrete labels from the region-based configuration's trained probabilities is described in Section~\ref{sec:region-order}.

\subsection{Training Configurations}
All nnU-Net-based models (baseline, ResEncL, and Primus) used nnU-Net's self-configured preprocessing and default optimization protocol, with patch size and batch size determined automatically per configuration via nnU-Net's fingerprinting step, and deep supervision enabled by
default. Plain nnU-Net and ResEncL used the default \texttt{nnUNetTrainer} (SGD, Nesterov momentum 0.99, initial LR~0.01 with polynomial decay, per-sample rather than batch Dice); Primus used its dedicated \texttt{nnUNet\_PrimusV3S\_Trainer} with that trainer's default settings. Inference used sliding-window prediction with mirroring test-time augmentation, with region probabilities binarized at 0.5. The plain nnU-Net baseline was trained for 1{,}000~epochs across 5~folds on an NVIDIA~A100 GPU; the earlier, shorter-trained checkpoint referenced above was trained for 200~epochs. ResEncL, region-based ResEncL, and both Primus configurations were each trained for 5~folds on NVIDIA~H100~SXM GPUs, with folds trained in parallel. All models were trained and evaluated on the identical patient-grouped 5-fold splits described above.

\subsection{Region-Order Assumption in Region-Based Label Reconstruction}
\label{sec:region-order}
Region-based nnU-Net reconstructs a discrete label map from overlapping region predictions by processing regions in a declared order, with each region overwriting prior assignments where they overlap. This assumes the declared order matches the regions' anatomical containment hierarchy, outermost to innermost. Our labeling scheme violates this under naive
ascending order: label~1 (NETC) is innermost, while label~2 (SNFH) is outermost, the reverse of what ascending order assumes. As a result, the whole-tumor region was assigned label~1 instead of label~2, inflating NETC with misclassified SNFH voxels and collapsing the tumor-core (TC) score. 
We identified this via a voxel-count comparison against the label-based model's output; the correct order applies the outer region first, then the inner tumor-core region, then the remaining independent classes.

\subsection{Postprocessing}
\label{sec:postprocessing}
Initial postprocessing applied morphological opening per class, which caused lesion-wise F1 to collapse to near zero (0.01--0.02) despite reasonable DSC, since the erosion step eliminates connected components near the 27~mm$^3$ scoring threshold before dilation can restore them. We removed morphological opening and replaced it with connected-component filtering on physical volume: components below 27~mm$^3$ are discarded, plus per-class hole-filling for the ensemble. This fix alone recovered lesion-wise F1 from~$\sim$0.01 to~$\sim$0.33--0.41 across ET/TC/WT. We highlight this failure mode because morphological opening is a common postprocessing default, and its interaction with small-lesion scoring thresholds is rarely reported explicitly in prior BraTS work.

\section{Results}
\label{sec:results}
Table~\ref{tab:results} reports lesion-wise DSC and NSD, per class, for all trained configurations on the official validation leaderboard. Table~\ref{tab:small_f1} reports lesion-wise F1, taken directly from the official scorer's small-instance F1 output for each submission.
\vspace{-1em}

\begin{table}[h]
\centering
\caption{Lesion-wise DSC / NSD on the validation set, by class.}
\label{tab:results}
\begin{tabular}{lcccc}
\hline
Model & ET & TC & WT & RC \\
\hline
Plain nnU-Net, 200~ep & 0.646 / 0.698 & 0.666 / 0.683 & 0.609 / 0.597 & 0.406 / 0.279 \\
Plain nnU-Net, 1000~ep & 0.664 / 0.727 & 0.688 / 0.737 & 0.652 / 0.675 & 0.478 / 0.376 \\
ResEncL, label-based & 0.693 / 0.755 & 0.708 / 0.757 & 0.676 / 0.699 & 0.415 / 0.345 \\
Primus, label-based & 0.710 / 0.761 & \textbf{0.742 / 0.785} & 0.683 / 0.689 & 0.531 / \textbf{0.436} \\
Primus, region-based & \textbf{0.713 / 0.762} & 0.400 / 0.433 & 0.680 / 0.682 & \textbf{0.566} / 0.376 \\
Ensemble (Primus + ResEncL) & 0.711 / 0.758 & 0.742 / 0.782 & \textbf{0.685 / 0.691} & 0.520 / 0.425 \\
\hline
\end{tabular}
\end{table}

\vspace{-3em}

\begin{table}[h]
\centering
\caption{Lesion-wise F1 on the validation set, by class.}
\label{tab:small_f1}
\begin{tabular}{lcccc}
\hline
Model & ET & TC & WT & RC \\
\hline
Plain nnU-Net, 200~ep & 0.012 & 0.012 & 0.002 & 0 \\
Plain nnU-Net, 1000~ep & 0.414 & 0.423 & 0.331 & 0 \\
\textbf{ResEncL, label-based} & \textbf{0.452} & \textbf{0.449} & \textbf{0.354} & 0 \\
Primus, label-based & 0.052 & 0.053 & 0.044 & 0 \\
Primus, region-based & 0.051 & 0.044 & 0.044 & 0 \\
Ensemble (Primus + ResEncL) & 0.064 & 0.065 & 0.044 & 0 \\
\hline
\end{tabular}
\end{table}

\vspace{-1em}
Primus (label-based) achieves the highest mean DSC/NSD across classes among individual models, leads on TC, and is the model used in our final ensemble; per-class differences among the top configurations are otherwise narrow, with the ensemble marginally ahead on WT and region-based Primus ahead on RC DSC specifically. Region-based Primus improves ET DSC/NSD (0.713/0.762 vs.~0.710/0.761) but underperforms on TC and RC NSD, a trade-off we consider a current limitation of the region-based approach. On lesion-wise F1 (Table~\ref{tab:small_f1}), both Primus configurations perform similarly poorly (0.044--0.053) and are substantially outperformed by ResEncL (0.354--0.452) on ET, TC, and WT; all models score zero on RC (discussed below).


\begin{figure}[H]
\centering
\includegraphics[width=\textwidth]{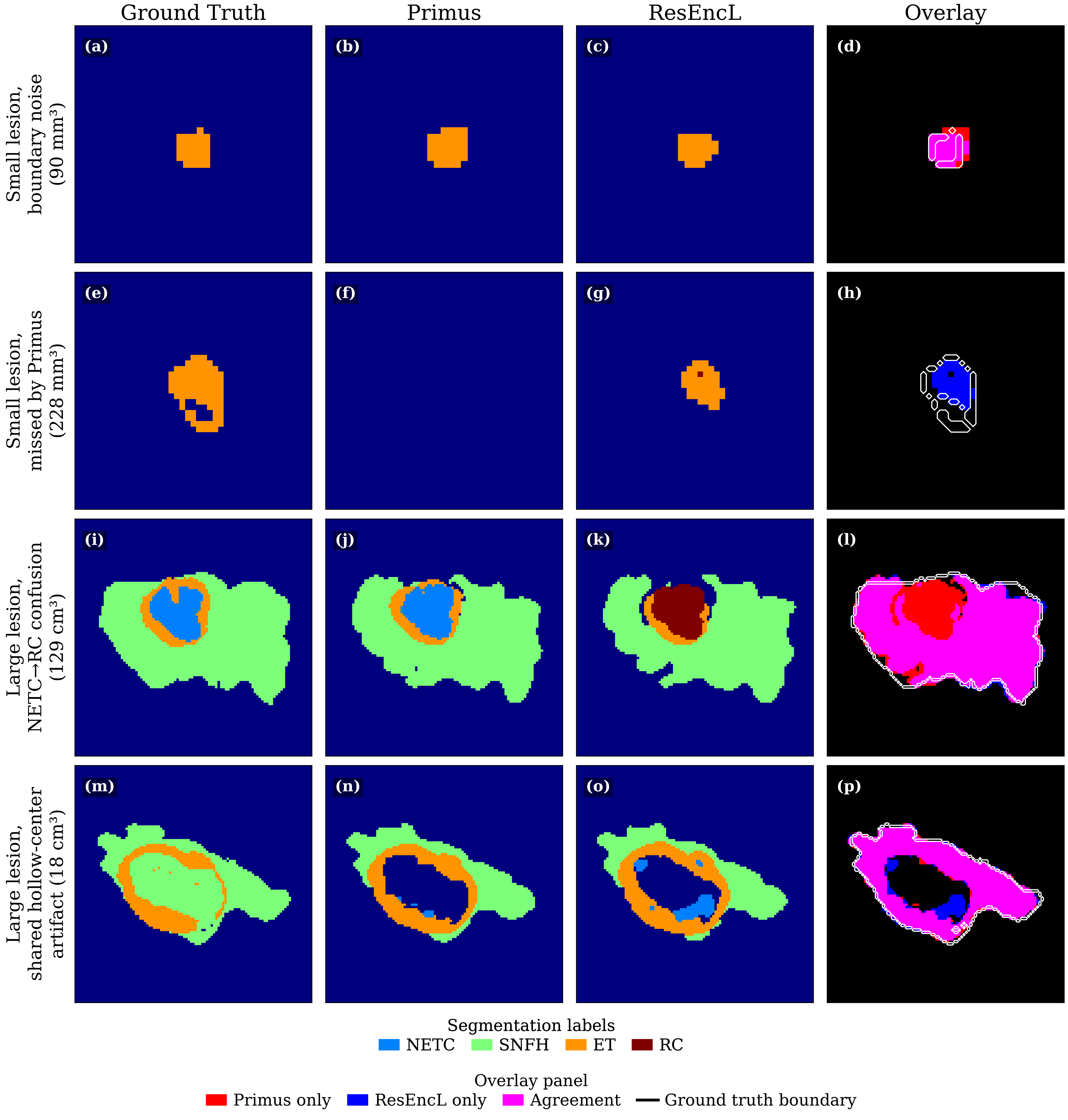}
\caption{Qualitative comparison of ResEncL and Primus segmentations against
ground truth for four representative cases, illustrating (row~1)
boundary-level disagreement on a small, correctly-detected lesion;
(row~2) a complete detection failure by Primus on a small lesion
correctly identified by ResEncL; (row~3) a case-specific substitution of
necrotic tumor core (NETC) with resection cavity (RC) by ResEncL; and
(row~4) a hollow-center prediction artifact shared by both models on a
ring-shaped lesion. Columns show ground truth, Primus prediction, ResEncL
prediction, and an overlay of the whole-tumor region
(NETC~$\cup$~SNFH~$\cup$~ET) where red indicates Primus-only, blue
indicates ResEncL-only, magenta indicates agreement, and the white
contour marks the ground-truth boundary.}
\label{fig:qualitative}
\end{figure}
\vspace{-1em}

Lesion-wise F1 reveals a starkly different pattern from DSC/NSD. ResEncL achieves the strongest lesion-wise F1 across ET/TC/WT (0.354--0.452), moderately exceeding even the nnU-Net baseline (0.331--0.423), while Primus lags far behind both (0.044--0.053) despite leading on aggregate overlap metrics. This dissociation between overlap-based accuracy and small-lesion detection is the central empirical
finding of this comparison: a model's DSC/NSD ranking is not predictive of its small-instance detection ranking, and the two must be evaluated independently. We attribute Primus's small-instance weakness to its transformer backbone's coarse 3D tokenization, which operates at a spatial scale close to that of the smallest scored lesions, while ResEncL's fully convolutional design preserves fine-grained spatial resolution better suited to sub-token-scale structures. This complementary strength profile suggests genuine architectural diversity between the two models, making them promising candidates for ensembling.

Figure~\ref{fig:qualitative} illustrates this dissociation qualitatively across four representative cases spanning small and large lesion scales. For small lesions, model disagreement is not uniform: in one case (90~mm$^3$), both models correctly localize and roughly match the lesion's extent, with the discrepancy confined to a small number of boundary voxels. In another case (228~mm$^3$), however, Primus fails to detect the lesion entirely while ResEncL correctly identifies it, indicating a genuine detection failure rather than boundary-level noise, consistent with the quantitative lesion-wise F1 gap above.

For large lesions, model behavior diverges differently. In one case (129~cm$^3$), ResEncL substitutes a substantial portion of the necrotic tumor core (NETC) with resection cavity (RC), a clinically distinct structure, while Primus's prediction closely matches the ground-truth NETC/ET layering; we observe this substitution in this case only and report it as a case-specific failure rather than a systematic tendency. A
separate case (18~cm$^3$) illustrates a shared limitation: both models under-predict the interior enhancing/necrotic composition of a ring-shaped lesion, producing a hollow region not present in the ground
truth.

\section{Discussion}
\label{sec:discussion}
The most impactful finding is that morphological postprocessing meant to clean up predictions can suppress detection of the small lesions the evaluation penalizes most heavily. The lesion-wise F1 metric is highly sensitive to whether small components survive postprocessing, so practitioners should check postprocessing effects on the small-lesion subset specifically, not just aggregate DSC.

A related pattern emerges from comparing Primus and ResEncL under identical postprocessing: Primus leads on DSC/NSD, but ResEncL's lesion-wise F1 is roughly an order of magnitude higher (e.g.\ ET: 0.052 vs.\ 0.452). Strong overlap metrics do not guarantee good small-lesion detection, and the two should be evaluated separately. Since ResEncL achieved strong DSC without losing lesion-wise F1, the two models appeared complementary, motivating our ensemble. In practice,
the ensemble's lesion-wise F1 (e.g.\ ET: 0.064) stayed close to Primus's rather than ResEncL's, despite weighting intended to protect ResEncL's contribution. This suggests probability-averaging ensembles do not automatically preserve a component model's minority-case strengths: when Primus assigns near-zero probability to a lesion it misses, averaging with ResEncL's moderate probability often still falls short of the 0.5 decision threshold. Threshold- or size-adaptive ensemble weighting may address this.

ResEncL also shows a distinct weakness on RC (DSC~0.415, the lowest of the individually-trained models), despite sitting between the nnU-Net baseline and Primus elsewhere. RC is a single, well-defined cavity rather than a small multifocal structure, so this is likely a separate, unexplained effect rather than the same cause as Primus's small-lesion gap.

The under-detection pattern (false negatives outweighing false
positives) seen in cross-validation points toward foreground
oversampling, false-negative-weighted losses such as Tversky loss, or a two-stage cascade as likely next steps. Given Primus's shortfall sits in the same small-lesion regime, we consider a two-stage cascade the highest-priority
next experiment.

RC remains a weakness across our pipeline. ResEncL performs worst (DSC~0.415), the nnU-Net baseline partially compensates (0.478), and region-based Primus achieves the highest RC DSC among our models (0.566), though label-based Primus leads RC NSD (0.436); even the best of these falls well short of competitive pipelines. Official small-instance scoring shows zero true positives for small resection-cavity instances ($\sim$1.17 per case, averaged) across every model we submitted -- nnU-Net, ResEncL, both Primus configurations, and the ensemble, indicating a small-RC detection failure that is universal across architectures rather than specific to any one model.

Our region-based training pipeline surfaces a labeling-order
assumption in nnU-Net's region reconstruction (Section~\ref{sec:region-order}) likely to recur in other setups where label containment order does not match ascending label values.

\noindent\textbf{Open items.} (1)~NSD tolerance in our offline evaluator is unverified against the official scoring script. (2)~The lesion matching rule is unconfirmed against the official evaluator. (3)~Missing-T2w validation cases are excluded at inference, which may cause silent score drag.

\section{Conclusion}

We compared four segmentation configurations for BraTS-METS~2026 Task~1, trained with patient-grouped CV and evaluated on the official validation leaderboard, finding that Primus and ResEncL showed complementary strongholds: Primus led on aggregate DSC/NSD, while ResEncL led substantially on lesion-wise F1, with neither model unambiguously best across both. Our probability-averaging ensemble only partially captured this complementarity, suggesting a more targeted ensembling strategy e.g.\ size- or threshold-adaptive weighting could recover more of ResEncL's small-lesion strength without sacrificing Primus's overlap accuracy. Future work includes such adaptive ensembling, a two-stage cascade for the small-lesion regime, and anatomical hierarchy enforcement for RC detection.

\subsubsection*{Acknowledgments}
Data used in this publication were obtained as part of the Challenge project. The authors gratefully acknowledge the BraTS "Fast-track-to-MICCAI" Educational Program for providing the computational resources and educational materials that supported this work. The program aims to provide students or early-career trainees, selected through a rigorous selection process, with resources for developing state-of-the-art solutions for brain tumor segmentation and detection on Magnetic Resonance Imaging (MRI) studies. This educational program is led by Mariam S. Aboian MD, PhD (Assistant Professor of Radiology at the Department of Radiology, Children's Hospital of Philadelphia) and is funded by the RSNA Derek Harwood-Nash International Education Scholar Grant. Beyond providing these resources, Dr. Aboian and her research team had no role in the conception of the research idea, algorithm development, data analysis, or preparation of this manuscript.

\subsubsection*{Disclosure of Interests}
The authors have no competing interests to declare that are relevant to the
content of this article.

\bibliographystyle{splncs04}
\bibliography{brats2026_refs}

@article{b_prevalence1,
  author  = {Lamba, Nayan and Wen, Patrick Y. and Aizer, Ayal A.},
  title   = {Epidemiology of brain metastases and leptomeningeal disease},
  journal = {Neuro-Oncology},
  volume  = {23},
  number  = {9},
  pages   = {1447--1456},
  year    = {2021},
  doi     = {10.1093/neuonc/noab101}
}

@incollection{b_prevalence2,
  author    = {Ostrom, Quinn T. and Wright, Carol H. and Barnholtz-Sloan, Jill S.},
  title     = {Brain metastases: Epidemiology},
  booktitle = {Handbook of Clinical Neurology},
  volume    = {149},
  pages     = {27--42},
  publisher = {Elsevier},
  year      = {2018},
  doi       = {10.1016/B978-0-12-811161-1.00002-5}
}

@article{b_survival,
  author  = {Yri, Odd E. and Astrup, Guro L. and Karlsson, Anne T. and Van Helvoirt, Robert and Hjermstad, Marianne J. and Husby, Kine M. and Loge, Jon H. and Lund, Jo-{\AA}ke and Lundeby, Tonje and Paulsen, {\O}yvind and Skovlund, Eva and Taran, Marte-Ida and Winther, Ragnhild R. and Aass, Nina and Kaasa, Stein},
  title   = {Survival and quality of life after first-time diagnosis of brain metastases: A multicenter, prospective, observational study},
  journal = {The Lancet Regional Health - Europe},
  volume  = {49},
  pages   = {101181},
  year    = {2025},
  doi     = {10.1016/j.lanepe.2024.101181}
}

@article{b_cost,
  author  = {Shim, Yu-Bin and Byun, Ji-Young and Lee, Ju-Young and Lee, Eui-Kyung and Park, Min-Hye},
  title   = {Economic burden of brain metastases in non-small cell lung cancer patients in {S}outh {K}orea: A retrospective cohort study using nationwide claims data},
  journal = {PLOS ONE},
  volume  = {17},
  number  = {9},
  pages   = {e0274876},
  year    = {2022},
  doi     = {10.1371/journal.pone.0274876}
}

@article{b_treatment,
  author  = {Nahed, Brian V. and Alvarez-Breckenridge, Christopher and Brastianos, Priscilla K. and Shih, Helen and Sloan, Andrew and Ammirati, Mario and Kuo, John S. and Ryken, Timothy C. and Kalkanis, Steven N. and Olson, Jeffrey J.},
  title   = {Congress of Neurological Surgeons Systematic Review and Evidence-Based Guidelines on the Role of Surgery in the Management of Adults With Metastatic Brain Tumors},
  journal = {Neurosurgery},
  volume  = {84},
  number  = {3},
  pages   = {E152--E155},
  year    = {2019},
  doi     = {10.1093/neuros/nyy542}
}

@misc{b_brats23,
  title         = {The Brain Tumor Segmentation (BraTS-METS) Challenge 2023: Brain Metastasis Segmentation on Pre-treatment MRI},
  author        = {Moawad, Ahmed W. and others},
  year          = {2024},
  eprint        = {2306.00838},
  archivePrefix = {arXiv},
  primaryClass  = {q-bio.OT},
  url           = {https://arxiv.org/abs/2306.00838}
}

@article{b_nnunet,
  author  = {Isensee, Fabian and Jaeger, Paul F. and Kohl, Simon A. A. and Petersen, Jens and Maier-Hein, Klaus H.},
  title   = {{nnU-Net: a self-configuring method for deep learning-based biomedical image segmentation}},
  journal = {Nature Methods},
  volume  = {18},
  pages   = {203--211},
  year    = {2021},
  doi     = {10.1038/s41592-020-01008-z}
}

@misc{b_nnunet_revisited,
  title         = {nnU-Net Revisited: A Call for Rigorous Validation in 3D Medical Image Segmentation},
  author        = {Isensee, Fabian and Wald, Tassilo and Ulrich, Constantin and Baumgartner, Michael and Roy, Saikat and Maier-Hein, Klaus and Jaeger, Paul F.},
  year          = {2024},
  eprint        = {2404.09556},
  archivePrefix = {arXiv},
  primaryClass  = {cs.CV},
  url           = {https://arxiv.org/abs/2404.09556}
}

@inproceedings{b_nnunet_brats,
  author    = {Isensee, Fabian and others},
  title     = {{nnU-Net for Brain Tumor Segmentation}},
  booktitle = {Brainlesion: Glioma, Multiple Sclerosis, Stroke and Traumatic Brain Injuries (BrainLes 2020)},
  series    = {Lecture Notes in Computer Science},
  volume    = {12659},
  publisher = {Springer, Cham},
  year      = {2021},
  doi       = {10.1007/978-3-030-72087-2_11}
}

@inproceedings{b_swinunetr,
  author    = {Hatamizadeh, Ali and others},
  title     = {{Swin UNETR: Swin Transformers for Semantic Segmentation of Brain Tumors in MRI Images}},
  booktitle = {Brainlesion: Glioma, Multiple Sclerosis, Stroke and Traumatic Brain Injuries (BrainLes 2021)},
  series    = {Lecture Notes in Computer Science},
  volume    = {12962},
  publisher = {Springer, Cham},
  year      = {2022},
  doi       = {10.1007/978-3-031-08999-2_22}
}

@misc{b_primus,
  title         = {Primus: Enforcing Attention Usage for 3D Medical Image Segmentation},
  author        = {Wald, Tassilo and Roy, Saikat and Isensee, Fabian and Ulrich, Constantin and Ziegler, Sebastian and Trofimova, Dasha and Stock, Raphael and Baumgartner, Michael and K{\"o}hler, Gregor and Maier-Hein, Klaus},
  year          = {2026},
  eprint        = {2503.01835},
  archivePrefix = {arXiv},
  primaryClass  = {cs.CV},
  url           = {https://arxiv.org/abs/2503.01835}
}

@InProceedings{b_astaraki,
  author    = {Astaraki, Mehdi and Moghadam, Farangis Sajadi and Toma-Dasu, Iuliana},
  title     = {Brain Tissue Context for Enhancing Brain Tumor Segmentation: A Contribution to BraTS 2025},
  booktitle = {Segmentation, Classification, and Synthesis for Brain Tumors and Traumatic Brain Injuries},
  year      = {2026},
  publisher = {Springer Nature Switzerland},
  address   = {Cham},
  pages     = {112--126},
  isbn      = {978-3-032-16365-3},
  doi       = {10.1007/978-3-032-16365-3_10}
}

@InProceedings{b_krikorian,
  author    = {Krikorian, Wes and Purwar, Ananya},
  title     = {Automated Segmentation for the Brain Tumor Segmentation (BraTS) Metastases 2025 Challenge Using Multi-Architectural Deep Learning},
  booktitle = {Segmentation, Classification, and Synthesis for Brain Tumors and Traumatic Brain Injuries},
  year      = {2026},
  publisher = {Springer Nature Switzerland},
  address   = {Cham},
  pages     = {173--181},
  isbn      = {978-3-032-16365-3},
  doi       = {10.1007/978-3-032-16365-3_16}
}

@article{b_brats25_lighthouse,
  author  = {Maleki, Nazanin and Amiruddin, Raisa and Moawad, Ahmed W. and Yordanov, Nikolay and Gkampenis, Athanasios and Fehringer, Pascal and Umeh, Fabian and Chukwurah, Crystal and Memon, Fatima and Petrovic, Bojan and others},
  title   = {{Analysis of the MICCAI Brain Tumor Segmentation -- Metastases (BraTS-METS) 2025 Lighthouse Challenge: Brain Metastasis Segmentation on Pre- and Post-treatment MRI}},
  year    = {2025},
  url = {https://arxiv.org/abs/2504.12527v3},
  
}

\end{document}